%% file: main.tex
\documentclass{article}

\usepackage{graphicx}
\usepackage{caption}
\usepackage[accsupp]{axessibility}  % Improves PDF readability for those with disabilities.
\usepackage{enumitem}
\usepackage{booktabs}     % 提供 \toprule, \midrule, \bottomrule
\usepackage{multirow}     % 提供 \multirow
\usepackage{adjustbox}    % 提供 adjustbox 环境
\usepackage[table]{xcolor} % 提供 \rowcolor (灰色背景)
\usepackage{amssymb}      % 提供 \checkmark (钩钩符号)
\usepackage{makecell}      % 必须有，解决 \makecell 报错
\usepackage[table]{xcolor} % 解决表格颜色
\usepackage[linesnumbered,ruled,vlined]{algorithm2e}
\usepackage[numbers,sort&compress]{natbib}

\usepackage[preprint]{neurips_2026}

\usepackage{hyperref}

\usepackage{orcidlink}
\usepackage{microtype}

\usepackage[utf8]{inputenc} % allow utf-8 input
\usepackage[T1]{fontenc}    % use 8-bit T1 fonts
\usepackage{hyperref}       % hyperlinks
\usepackage{url}            % simple URL typesetting
\usepackage{booktabs}       % professional-quality tables
\usepackage{amsfonts}       % blackboard math symbols
\usepackage{nicefrac}       % compact symbols for 1/2, etc.
\usepackage{microtype}      % microtypography
\usepackage{xcolor}         % colors
\usepackage{caption}
\usepackage{cleveref}

\title{KeyframeFace: Language-Driven Facial Animation via Semantic Keyframes}
\author{%
  Jingchao Wu$^{1,2}$\thanks{Equal contribution.} \quad
  Zejian Kang$^{3,1}$\footnotemark[1] \quad
  Haibo Liu$^{1}$ \quad
  Yuanchen Fei$^{1,4}$ \quad
  Xiangru Huang$^{1}$\thanks{Corresponding author.} \\[0.5em]
  $^1$Westlake University \quad
  $^2$Nanjing University \quad
  $^3$Zhejiang University \quad
  $^4$Hunan University
}

\begin{document}
\maketitle

\begin{center}
    \includegraphics[width=1\linewidth]{picture/teaser_new.jpg}
    \captionof{figure}{
        \textbf{Facial animation generated by KeyframeFace.} Given descriptions of contextual background and desired emotions, our text-to-animation framework produces keyframe-level facial descriptions and generates the corresponding ARKit coefficients that can be directly converted into expressive facial animations and videos via tools like MetaHuman~\cite{metahuman}.
        }
    \label{fig:teaser}
\end{center}
% \vspace{-0.5cm}

\input{sec/0_abstract}   
\input{sec/1_intro}  
\input{sec/2_related_work}

\input{sec/3_dataset}

\input{sec/4_method}
\input{sec/5_exp}
\input{sec/6_conclusion}

\clearpage
\newpage
\bibliographystyle{abbrvnat}  %plainnat,abbrvnat,unsrtnat
\bibliography{main}

%%%%%%%%%%%%%%%%%%%%%%%%%%%%%%%%%%%%%%%%%%%%%%%%%%%%%%%%%%%%
\clearpage
\appendix

\input{sec/appendix}

%%%%%%%%%%%%%%%%%%%%%%%%%%%%%%%%%%%%%%%%%%%%%%%%%%%%%%%%%%%%
% \clearpage
% \newpage
% \input{checklist.tex}

\end{document}

%% file: sec/0_abstract.tex
\begin{abstract}

Facial animation is a core component for creating digital characters in Computer Graphics (CG) industry. A typical production workflow relies on sparse, semantically meaningful keyframes to precisely control facial expressions. Enabling such animation directly from natural-language descriptions could significantly improve content creation efficiency and accessibility. However, most existing methods adopt a text-to-continuous-frames paradigm, directly regressing dense facial motion trajectories from language. This formulation entangles high-level semantic intent with low-level motion, lacks explicit semantic control structure, and limits precise editing and interpretability. Inspired by the keyframe paradigm in animation production, we propose KeyframeFace, a framework for semantic facial animation from language via interpretable keyframes. Instead of predicting dense motion trajectories, our method represents animation as a sequence of semantically meaningful keyframes in an interpretable ARKit-based facial control space. A language-driven model leverages large language model (LLM) priors to generate keyframes that align with contextual text descriptions and emotion cues. To support this formulation, we construct a multimodal dataset comprising 2,100 expression scripts paired with monocular videos, per-frame ARKit coefficients, and manually annotated semantic keyframes. Experiments show that incorporating semantic keyframe supervision and language priors significantly improves expression fidelity and semantic alignment compared to methods that do not use facial action semantics.
% \keywords{Facial Animation \and Large Language Model \and Multimodal Dataset \and Text-to-animation}

\end{abstract}

%% file: sec/1_intro.tex
\section{Introduction}
\label{sec:intro}

Facial animation is a long-standing challenge in the CG industry, with broad applications in film production, gaming, and virtual avatars. In professional animation pipelines, facial performance is typically constructed through a keyframe-driven workflow, where animators define a sparse set of key expression states at semantically critical moments, and intermediate motion is obtained via interpolation. This formulation provides explicit control over expression timing, intensity, and transitions, and is widely adopted in industry practice.

Recent work has explored generating facial animation from natural-language descriptions~\cite{ma2025talkclip,zhong2024expclip,zhao2024media2face,wu2024mmhead,aloni2025express4d,xu2025words, jiang2025emocast}. These methods typically adopt a text-to-continuous-frames paradigm, directly regressing dense per-frame facial parameters or latent expression trajectories from text input. While such approaches can produce plausible motion sequences, they entangle high-level semantic intent with low-level motion details, offering limited interpretability and controllability. As a result, it remains difficult for users to precisely specify or edit key expressive moments, such as the onset of a smile or the timing of an eyebrow raise, which are essential for professional-quality animation.

An effective strategy to address these limitations is to formulate the problem at the level of semantic keyframes rather than dense trajectories, mirroring how professional animators work in practice. This shifts the modeling target from low-level per-frame motion to high-level expression semantics, yielding interpretable and editable control over expression timing and intensity. Beyond this structural advantage, LLMs encode rich priors about emotions and expression semantics that can be harnessed to further improve contextual understanding in language-driven animation.

Building on these insights, we introduce \textbf{KeyframeFace}, a framework for language-driven facial animation via interpretable keyframes. Specifically, our method represents facial animation as a sparse sequence of keyframes in the ARKit blendshape space~\cite{arkit}, and leverages LLM priors to generate these keyframes from textual context and emotion cues. Accordingly, to support this formulation, we construct a large-scale multimodal dataset comprising 2,100 expressive scripts performed by human actors. Each keyframe is manually annotated with both ARKit coefficients and detailed textual descriptions, providing rich semantic supervision for training.

Our main contributions are summarized as follows:
\begin{itemize}

    \item We introduce a \textbf{keyframe-centric formulation} for facial animation that represents expressions as semantic keyframes rather than dense motion trajectories, bridging the gap between high-level language semantics and low-level facial motion.
    
    \item We present \textbf{KeyframeFace}, a text-to-facial-animation framework that generates semantic keyframes in an interpretable ARKit blendshape space, enabling precise and editable control over expression timing and intensity.
    
    \item We construct \textbf{a large-scale multimodal dataset} with script-level semantic supervision and manually annotated expression keyframes.
    
    \item We demonstrate that \textbf{LLM priors} on emotions and expression semantics can be effectively applied to facial animation, providing contextual understanding for keyframe generation.
\end{itemize}

%% file: sec/2_related_work.tex
\section{Related Work}
\label{sec:relatedwork}

\paragraph{Text-Driven Facial Animation.}
Recent advances in facial animation have demonstrated impressive results in expressive performances from both audio and language inputs. While audio-driven methods have matured significantly~\cite{fan2022faceformer, richard2021meshtalk, sun2024diffposetalk, ma2024diffspeaker,xing2023codetalker,stan2023facediffuser,zhang2026ex,bigata2025keyface,chung2025audio2face}, text-driven approaches offer a more accessible and semantically expressive control modality~\cite{ma2025talkclip, zhong2024expclip}. Most existing text-driven methods rely on 3DMM/FLAME parameters or mesh displacements~\cite{ma2025talkclip, zhong2024expclip}, which encode geometric deformations as basis-shape combinations with no inherent correspondence to facial actions. In contrast, ARKit's 61 blendshape coefficients, introduced with iPhone’s TrueDepth sensor, are explicitly named after anatomically grounded facial actions (e.g., \textit{browInnerUp}, \textit{jawOpen}), enabling direct alignment between natural language and motion parameters~\cite{chung2025audio2face, stan2023facediffuser}. However, limited research has combined LLM semantic reasoning with ARKit's parameter space for text-driven animation—--an approach we pursue in this work.

For temporal modeling, frame-level autoregressive approaches~\cite{fan2022faceformer,xing2023codetalker, chu2025artalk} achieve strong synchronization but accumulate errors over long sequences, while multi-track timeline control~\cite{ma2025exploring} enables precise scheduling yet demands dense manual annotations. More recently, KeyFace~\cite{bigata2025keyface} offers a more structured alternative by anchoring generation at expressive moments and interpolating intermediate transitions, but lacks direct text-conditioned control. To this end, we propose a unified keyframe-based framework for text-driven 3D facial animation.

% \vspace{0.5em}
\paragraph{3D Facial Motion Datasets.} 
Existing 3D facial motion datasets vary widely in modality composition and annotation granularity. Early datasets~\cite{zhang2008high,zhang2016multimodal,cheng20184dfab,matuszewski2012hi4d} mainly include basic categorical affect labels over short clips, offering limited supervision for modeling fine-grained temporal dynamics. Audio-driven corpora~\cite{cudeiro2019capture,fanelli20103,wang2020mead,zhang2021flow,wu2023mmface4d,he2023speech4mesh,peng2023emotalk} provide temporally coherent 3D facial signals but focus on speech synchronization. Only a few studies have attempted to incorporate natural language descriptions into facial motion generation. MMHead~\cite{wu2024mmhead} automatically generates textual descriptions from video content, but the annotation quality is unstable. Express4D~\cite{aloni2025express4d} drives expressions with free-text prompts, linking text to facial motion, but its short and simple prompts limit the modeling of complex or multi-stage dynamics. \Cref{tab:dataset_comparison} provides a systematic comparison of representative facial expression datasets.
Compared with existing works, our proposed dataset achieves notable improvements in script complexity, emotional richness, and multimodal data integration.

%% file: sec/3_dataset.tex
\section{KeyframeFace Dataset}
\label{sec:dataset}

\begin{figure*}[h]
    \centering
   \makebox[\textwidth][c]{%
        \includegraphics[width=1.0\textwidth]{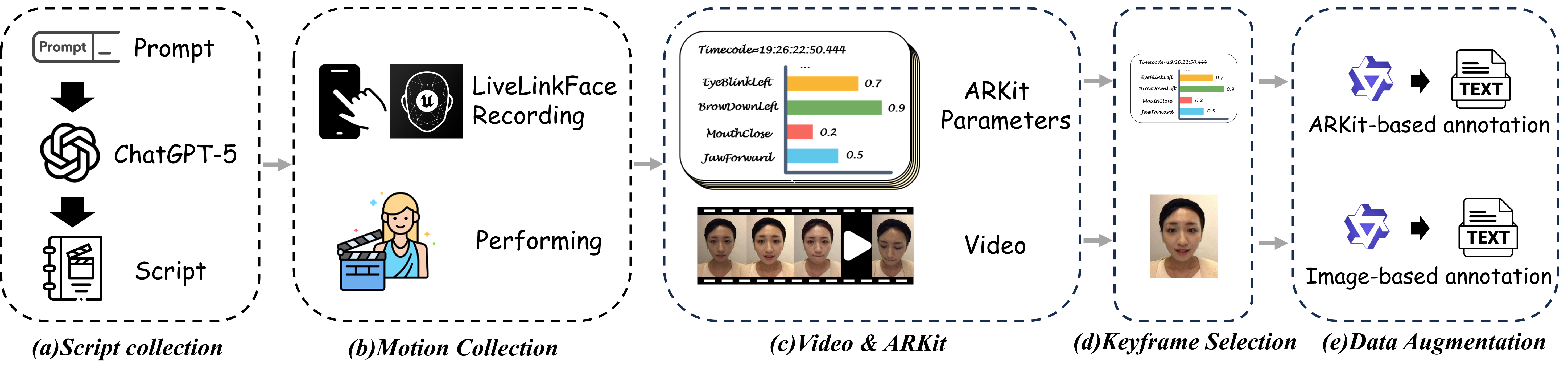}  % 115% 页宽
    }
    \caption{\textbf{Overview of our data pipeline.} (a) Script generation with contextual keyframe descriptions, (b) ARKit-based motion capture, (c) Synchronized video and coefficient recording, (d) Manual keyframe selection, and (e) Multi-perspective augmentation using LLMs and Multimodal Large Language Models (MLLMs).}
    \label{fig:data_pipeline}
\end{figure*}

We design a unified data pipeline to construct KeyframeFace Dataset, as illustrated in~\Cref{fig:data_pipeline}.
% To construct KeyframeFace Data, we designed a unified data pipeline that links high-level emotional scripts to fine-grained facial motion sequences (\cref{fig:data_pipeline}). 
Specifically, we detail the dataset construction process in \Cref{subsec:construction}, the multi-perspective data augmentation strategy in \Cref{subsec:data_augmentation}, and the comparison with existing datasets in \Cref{subsec:data_comparison}.

\subsection{Dataset Construction}
\label{subsec:construction}

\paragraph{Script Collection.}
% To construct KeyframeFace Data, we designed a unified data pipeline that links high-level emotional scripts to fine-grained facial motion sequences (\cref{fig:data_pipeline}). 
The process begins with Script Collection, where we used ChatGPT-5~\cite{openai2025gpt5} to generate 2,100 unique expressive scripts. To capture the subtle nuances of complex emotions, each script is designed to provide: (1)~rich situational background; (2)~composite emotional states; (3)~explicit keyframe-level descriptions.
For example, a prompt requesting a ``forced smile'' yields the following hierarchical structure: \begin{itemize} \item \textbf{\textit{Background:}} The subject is ill but does not want their family to worry. \item \textbf{\textit{Emotion:}} Suppressed Distress.\item \textbf{\textit{Keyframe 1:}} Eyes appear slightly fatigued; lips remain flat. \item \textbf{\textit{Keyframe 2:}} Mouth corners lift into a forced smile, while the gaze remains weak and faint wrinkles persist between the brows. \end{itemize}

\paragraph{Motion Collection.}
To ensure interpretability and seamless integration with production-grade animation pipelines, we adopt the ARKit blendshape format as our motion representation. Facial performances were captured at 60 Hz using an iPhone equipped with a TrueDepth camera, yielding 61-dimensional semantic blendshape vectors extracted per frame. Leveraging this high-fidelity yet accessible setup, we curated a dataset of 2,100 motion sequences from 21 professional actors. Each sequence was recorded following specific scripts to ensure realistic, context-aware facial expressions. Actors were provided with scripts covering diverse real-world 
scenarios and emotional contexts, and instructed to perform 
naturally, ensuring the collected expressions are both 
contextually grounded and expressively diverse.

% \vspace{0.5em}
\paragraph{Keyframe Selection.}
For each motion sequence, keyframes are manually selected based on 
the generated scripts, where each keyframe captures the peak of a 
specific emotional transition or microstate. Scripts are adjusted 
where necessary to reflect the actor's actual performance, ensuring 
consistency between textual descriptions and captured expressions. 
For every keyframe, we store the frame index, the corresponding 
ARKit coefficient vector, and the extracted frame image.

\subsection{Data Augmentation}
\label{subsec:data_augmentation}

The original keyframe descriptions mainly capture high-level semantic intent, while professional actors often introduce subtle facial nuances during performance. 
To enrich the supervision signal, we augment each keyframe annotation using both ARKit-based and image-based descriptions. 
For ARKit-based annotation, an LLM converts blendshape coefficients into natural facial-action descriptions, leveraging the interpretable correspondence between ARKit parameters and facial muscle activations. 
For image-based annotation, an MLLM describes fine-grained visual cues from the keyframe images, such as gaze changes, brow motion, eye tension, cheek deformation, and nasolabial folds.

All generated descriptions are manually reviewed to ensure that they faithfully reflect the captured actor performance while avoiding explicit ARKit coefficient values or script-specific terminology. 
This process provides a fine-grained and physically grounded linguistic layer that complements the macro-level intent in the original scripts. 
Further details on prompts, annotation examples, and quality control are provided in Appendix~\ref{subsec:data_augmentation_app}.

\subsection{Data Comparison}
\label{subsec:data_comparison}

\begin{table*}[h]
\centering
\small
\renewcommand{\arraystretch}{1.25}
\caption{Comparison of 3D facial motion datasets. 
``Scn.'', ``KF'', ``Emo.'', and ``Desc.'' denote scenario, keyframe, emotion, and facial description annotations, respectively.
``Dur.'' denotes duration. For ``Emo. Diversity'': numbers indicate fixed categories; $T$ denotes text-based vocabulary; $^{*}$ denotes mixed/composite emotion support. ``Mono.'' means monocular reconstruction.}
\label{tab:dataset_comparison}
\begin{adjustbox}{width=\textwidth,center}
\begin{tabular}{l
c
cc
c
cccc
c
c}
\toprule
\multirow{2}{*}{\textbf{Dataset}}
& \multirow{2}{*}{\textbf{Representation}}
& \multicolumn{2}{c}{\textbf{Modality}}
& \multicolumn{1}{c}{\textbf{Scale}}
& \multicolumn{4}{c}{\textbf{Annotation}}
& \multirow{2}{*}{\shortstack{\textbf{Emo.}\\\textbf{Diversity}}}
& \multirow{2}{*}{\textbf{Device}} \\
& & Video & Text
& Dur.(min)
& Scn. & KF & Emo. & Desc.
& & \\ 
\midrule
BU-4DFE~\cite{zhang2008high}& Mesh
& \checkmark & -
& 40
& - & - & \checkmark & -
& 6 & DI3D \\
D3D-FACS~\cite{cosker2011facs}& Mesh
& \checkmark & -
& 65
& - & \checkmark & - & -
& - & 3dMD \\
Hi4D-ADSIP~\cite{matuszewski2012hi4d}& Mesh
& \checkmark & -
& -
& - & - & \checkmark & -
& 7& DI3D \\
BP4D+~\cite{zhang2016multimodal}& Mesh
& \checkmark & -
& 777
& - & - & \checkmark & -
& 10& DI3D \\
4DFAB~\cite{cheng20184dfab}& Mesh
& \checkmark & -
& 509
& - & \checkmark & \checkmark & -
& 6& DI4D \\
VOCASET~\cite{cudeiro2019capture}& Mesh(FLAME)
& - & -
& 29
& - & - & - & -
& - & 3dMD \\
Florence4D~\cite{principi2023florence}& Mesh
& - & -
& 5075
& - & - & \checkmark & -
& 8 & 3dMD \\
MMFace4D~\cite{wu2023mmface4d}& Mesh(3DMM)
& \checkmark & -
& 2166
& - & - & \checkmark & -
& 7& RGB-D \\
MEAD-3D~\cite{he2023speech4mesh}& Mesh(FLAME)
& \checkmark & -
& 2280
& - & - & \checkmark & -
& 8 & Mono. \\
% \midrule
3D-ETF~\cite{peng2023emotalk}& 52Blendshapes(FLAME)
& \checkmark & -
& 390
& - & - & \checkmark & -
& 8 & Mono. \\
\midrule
MMHead~\cite{wu2024mmhead}& 56Blendshapes(FLAME)
& \checkmark & \checkmark
& 2940
& \checkmark & - & \checkmark & \checkmark
& $T$ & Mono. \\
Express4D~\cite{aloni2025express4d}& 61Blendshapes(ARKit)
& - & \checkmark
& 90
& - & - & - & \checkmark
& $T^{*}$ & Cellphone \\
\rowcolor{gray!10}
\textbf{KeyframeFace (Ours)}
& \textbf{61Blendshapes(ARKit)}
& \textbf{\checkmark} & \textbf{\checkmark}
& \textbf{330}
& \textbf{\checkmark} & \textbf{\checkmark} & \textbf{\checkmark} & \textbf{\checkmark}
& \textbf{$T^{*}$} & \textbf{Cellphone} \\
\bottomrule
\end{tabular}
\end{adjustbox}
\end{table*}

Our dataset offers significant advantages over existing 3D facial motion datasets. As summarized in~\Cref{tab:dataset_comparison}, KeyframeFace Dataset is the only dataset that jointly provides video input, text annotations, scenario labels, keyframe annotations, emotion labels, and detailed facial descriptions. Notably, it supports text-based complex and composite emotions, enabling richer affective representation compared to datasets with fixed emotion categories. Unlike existing lab-only 4D capture datasets that rely on professional equipment (e.g., DI3D, 3dMD), our data are recorded using commodity smartphones across diverse real-world environments, providing more natural expressions and greater ecological validity. Furthermore, our use of 61 ARKit blendshapes ensures strong interpretability and seamless compatibility with widely-used production pipelines such as MetaHuman. As ARKit blendshape coefficients encode muscle activation magnitudes independent of facial geometry, 21 actors performing 2,100 semantically diverse scripts provide sufficient coverage of the expressive motion space for identity-agnostic text-to-coefficient mapping. Additional visualizations and comprehensive dataset statistics are included in Appendix~\ref{appendix:data_v}.

%% file: sec/4_method.tex
\section{KeyframeFace Method}
\label{sec:Method}
\begin{figure*}[b!]
    \centering
   \makebox[\textwidth][c]{%
        \includegraphics[width=1.0\textwidth]{picture/method.jpg}  % 115% 页宽
    }
    \caption{\textbf{Overview of our KeyframeFace Framework.} Our approach involves: (a) Input Standardization stage that transforms diverse user inputs into unified keyframe descriptions through LLM-based analysis, and (b)  Text-To-Animation that generates and renders facial animations via a fine-tuned LLM.}
   
    \label{fig:pipeline}
\end{figure*}

In this section, we propose KeyframeFace, a two-stage framework that transforms arbitrary natural language descriptions into ARKit blendshape keyframe sequences (formalized in~\Cref{subsec:problem_formulation}).
As illustrated in~\Cref{fig:pipeline}, the first stage (Input Standardization,~\Cref{subsec:input_standardization}) converts diverse user inputs into unified keyframe descriptions. The second stage (Text-to-Animation,~\Cref{subsec:text_to_animation_model}) generates corresponding ARKit parameter sequences through 
fine-tuned LLMs based on keyframe descriptions. 

\subsection{Problem Formulation}
\label{subsec:problem_formulation}

Inspired by the keyframe paradigm in animation production, we reformulate text-to-facial-animation as sparse keyframe generation: given a user input $\mathcal{T}$ ranging from emotion words to detailed behavioral descriptions, our goal is to predict a structured action-value set $\mathcal{O} = \left\{ \left\{ (a_k, v_k^n) \right\}_{k=1}^{K} \right\}_{n=1}^{N}$. In this formulation, $N$ denotes the number of generated keyframes, dynamically dictated by the complexity of $\mathcal{T}$; $\mathcal{A} = \{a_k\}_{k=1}^{K}$ denotes the ARKit blendshape action set ($K = 61$, e.g., \textit{jawOpen}, \textit{mouthSmile}, \textit{browInnerUp}); and $v_k^n \in [-1, 1]$ specifies the activation value of the $k$-th blendshape at the $n$-th keyframe. This reformulation explicitly decouples high-level semantic intent from low-level motion, enabling interpretable and precisely controllable facial animation synthesis that prior approaches struggle to achieve.

\subsection{Input Standardization}
\label{subsec:input_standardization}
To handle the significant variability in user input granularity and format, we employ a pre-trained LLM to standardize $\mathcal{T}$ into a structured script $S$, extracting emotional context, temporal structure, and expression details into a consistent format. $S$ is presented to the user for confirmation and refinement before being passed to the next stage, ensuring full controllability over the animation intent. As this stage leverages the LLM's inherent semantic understanding capabilities, it generalizes to arbitrary input formats without additional training.

\subsection{Text-To-Animation Model}
\label{subsec:text_to_animation_model}
To bridge the gap between natural language semantics and facial 
motion parameters, we leverage the natural alignment between 
ARKit's anatomically grounded coefficients and the LLM's 
pre-trained knowledge of facial anatomy and expression, fine-tuning an LLM in the second stage to convert standardized keyframe descriptions into precise ARKit parameter sequences. This stage comprises three components: prompt engineering to align the model's understanding with ARKit parameter semantics, a recursive generation strategy for Text-To-ARKit conversion, and ARKit-To-Visual Animation rendering for high-fidelity facial animation synthesis.
% \vspace{-0.75em}
\paragraph{Prompt Engineering.} We design a structured system prompt comprising System Overview, 
Parameter Explanation, and Output Specification to guide the LLM 
in generating accurate ARKit parameters. Specifically, we construct a semantic mapping for all 61 ARKit coefficients, associating each parameter with its corresponding facial muscle activation pattern in textual form, enabling the model to leverage its pre-trained knowledge of facial anatomy for precise parameter generation.

\paragraph{Text-To-ARKit Generation.} As illustrated in~\Cref{alg:generation} in Appendix, we adopt a recursive keyframe generation strategy to ensure generation quality and temporal consistency. Given script $S$, we extract a sequence of keyframe descriptions $\{(F_n, n)\}_{n=1}^{N} \leftarrow f_{\text{key}}(S)$, enabling the handling of arbitrary-length sequences without predetermined constraints. For each keyframe description $F_n$, $f_{\text{task}}$ formats it into a target frame instruction $I_n$, which directs the model to focus on generating ARKit coefficients for the $n$-th keyframe (e.g., ``Please generate the facial expression parameters for Keyframe $n$: [$F_n$] and ...''). The LLM then receives a full input $U_n = P \oplus S \oplus I_n$ composed of the system prompt $P$, script $S$, and target frame instruction $I_n$, and generates the structured action-value set $\{(a_k, v_k^n)\}_{k=1}^{K}$ for the $n$-th keyframe. By iterating through all $N$ keyframes, we construct the complete output $\mathcal{O} = \left\{\{(a_k, v_k^n)\}_{k=1}^{K}\right\}_{n=1}^{N}$. By conditioning on the global context $S$ and specific frame instruction $I_n$, the model synthesizes parameters that are not only locally precise but also globally coherent. The LLM is fine-tuned with a standard causal language modeling objective:
\begin{equation}
\mathcal{L}_{\text{causal}} = - \frac{1}{L} \sum_{l=1}^{L} \log \frac{\exp(z_{l,x_l})}{\sum_{v \in V} \exp(z_{l,v})}
\end{equation}
where $L$ is the token sequence length, $z_{l,v}$ is the predicted logit for token $v$ at position $l$, $x_l$ is the ground-truth (GT) token, and $V$ is the vocabulary.

\paragraph{ARKit-To-Visual Animation.} The generated keyframe sequences $\mathcal{O}$ are transformed into high-fidelity visual animations via the Epic Games' MetaHuman system. To facilitate seamless transitions between the $N$ sparse keyframes, we propose an interpolation framework comprising two synergistic components:

\begin{itemize}
    \item \textbf{Duration Predictor $\mathcal{D}$:} This module predicts the intrinsic intermediate frame count $\hat{M}$ between adjacent keyframes vectors $\{\mathbf{v}_n, \mathbf{v}_{n+1}\} \subset \mathcal{O}$, where $\mathbf{v}_n = \{v_k^n\}_{k=1}^{K}$. To allow for flexible motion control, users can modulate this value via a speed scale factor $\alpha$, yielding the final target frame count $M = \lfloor \hat{M} \cdot \alpha \rceil$.
   \item \textbf{Motion Interpolator $\mathcal{F}$:} This component generates a canonical intermediate sequence $\{\mathbf{v}_{n,q}\}_{q=1}^Q \in \mathbb{R}^{Q \times K}$, which represents a normalized motion trajectory consisting of $Q$ frames conditioned on $\{\mathbf{v}^n, \mathbf{v}^{n+1}\}$.
\end{itemize}

At inference, the $Q$-frame sequence is resampled to length $M$ and concatenated with the original keyframes to construct the complete animation stream for rendering. 

%% file: sec/5_exp.tex
\section{Experiments}
\label{sec:experiments}

\subsection{Experimental Setup}

\paragraph{Baselines.} We compare our method against three baseline approaches, all retrained on our KeyframeFace dataset using identical data splits and ARKit-based motion representations to ensure fair comparison. Express4D-MDM~\cite{aloni2025express4d} adapts the Motion Diffusion Model (MDM)~\cite{tevet2022human} to facial animation, employing a diffusion-based framework for motion sequence generation. T2M-GPT~\cite{zhang2023generating} applies a VQ-VAE tokenization scheme with a GPT-based autoregressive prior, originally proposed for body motion generation and here adapted to the facial domain. CTEG~\cite{xu2025words} is a continuous text-to-expression generator based on an autoregressive Conditional Variational Autoencoder, designed specifically for emotionally coherent facial expression synthesis.

\paragraph{Evaluation Metrics.}
We evaluate our models using both a user study and several quantitative metrics.
The user study assesses the perceptual quality and preference of the generated facial animations via a Win Rate.
Mean Squared Error (MSE) and Mean Absolute Error (MAE) measure the numerical accuracy of predicted ARKit coefficients.
Wasserstein Distance (W-Dist) evaluates distributional alignment between predicted and ground-truth (GT) coefficient distributions.
R-Precision measures semantic consistency between facial actions and their textual descriptions, while Multimodal Distance (MMD) assesses cross-modal coherence in the learned embedding space.
Lower values indicate better performance for all metrics except R-Precision, where higher values reflect stronger semantic alignment.
Detailed formulations of these metrics are provided in Appendix~\ref{appendix:rprecision_mmd_training}.

\paragraph{Implementation Details.}
We conducted a comprehensive performance evaluation on KeyframeFace Dataset using three mainstream LLMs: Qwen3-4B-Instruct-2507
% (fine-tuning 17M out of 4039M parameters, 0.4089\%)
, Qwen3-14B
% (fine-tuning 32M out of 14800M parameters, 0.2170\%)
, and DeepSeek-R1-Distill-Qwen-14B\cite{guo2025deepseek}. 
% (fine-tuning 32M out of 14800M parameters, 0.2170\%)
All models were fine-tuned using LoRA\cite{hu2022lora} with $r = 8$ and $\alpha = 32$, targeting all linear layers including attention projections ($W_q$, $W_k$, $W_v$, $W_o$) and feed-forward networks. Training was conducted on $8\times\text{A100}$ GPUs for 90 epochs and implemented using the ms-SWIFT infrastructure\cite{zhao2025swift}.

\subsection{Comparison Results}
\begin{table}[h]
\centering
\caption{Quantitative comparison between KeyframeFace and baseline methods.}
\label{tab:quantitative_comparison}
\scriptsize
\setlength{\tabcolsep}{4pt}
\begin{tabular}{l cc c ccc c c}
\toprule
\multirow{2}{*}{\textbf{Setting}} 
& \multicolumn{2}{c}{\textbf{Coefficient Error}}
& \multicolumn{1}{c}{\textbf{Similarity}}
& \multicolumn{3}{c}{\textbf{R-Precision} $\uparrow$} 
& \multirow{2}{*}{\textbf{MMD} $\downarrow$}
& \multirow{2}{*}{\shortstack{\textbf{Win Rate} $\uparrow$\\\textbf{(Ours vs. Method)}}}\\
\cmidrule(lr){2-3} \cmidrule(lr){4-4} \cmidrule(lr){5-7}
& \textbf{MSE} $\downarrow$ & \textbf{MAE} $\downarrow$ 
& \textbf{W-Dist} $\downarrow$ 
& \textbf{top-1} & \textbf{top-2} & \textbf{top-3} & & \\
\midrule
\rowcolor{gray!15} \textbf{GT-test}
& - & - & 0.0055 & 0.2040 & 0.3565 & 0.4802 & 1.0600 & - \\
\midrule
CTEG
& \underline{0.0130}& \underline{0.0674}& 0.0501& 0.0397& 0.0661& 0.101& 1.4119& 72.4\% \\
T2M-GPT
& 0.0279& 0.1000& 0.0488& 0.0325& 0.0637& 0.0986& 1.4265& 91.7\% \\
Express4D-MDM
& 0.0229 & 0.0903 & \underline{0.0230} & \underline{0.0877} & \underline{0.1623} & \underline{0.2344} & \underline{1.2555} & 89.7\% \\
\textbf{KeyframeFace (Ours)}
& \textbf{0.0069}
& \textbf{0.0388} 
& \textbf{0.0145} 
& \textbf{0.2079} 
& \textbf{0.3413} 
& \textbf{0.4579} 
& \textbf{1.0602}
& - \\
\bottomrule
\end{tabular}
\end{table}

\paragraph{Quantitative comparison.}
As shown in~\Cref{tab:quantitative_comparison}, all three baselines exhibit significantly lower R-Precision scores and higher coefficient errors compared to our method. This indicates that fine-grained facial animation presents unique challenges: standard diffusion models often find it difficult to maintain explicit semantic grounding, discrete autoregressive models can be susceptible to prior collapse, and deterministic generation tends to produce overly smoothed results. By contrast, our method overcomes these structural bottlenecks and substantially outperforms all baselines across every metric, reducing MAE to 0.0388 and achieving an R-Precision top-1 of 0.2079. User study results further corroborate these findings, with our method achieving win rates of 72.4\%, 91.7\%, and 89.7\% against CTEG, T2M-GPT, and Express4D-MDM respectively. More discussion is included in Appendix~\ref{app:baseline_analysis}.

\paragraph{Visualization.}

\begin{figure}[h] % 建议使用 [t] (顶部) 或 [h] (此处)
    \centering
    \includegraphics[width=1.0\linewidth]{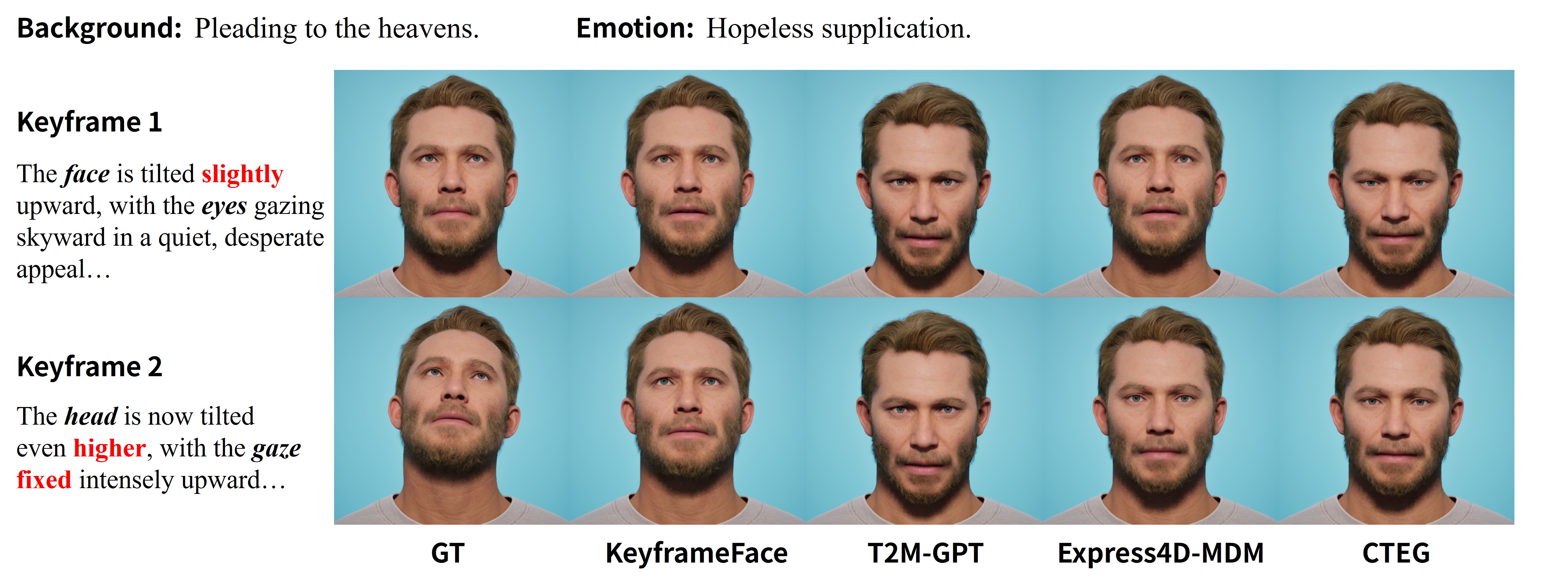} 
    \caption{\textbf{Generating Facial Expressions from Text: KeyframeFace vs. Baselines.}
    Our method faithfully captures the background context, emotional cues, and keyframe descriptions in the text prompt, enabling it to generate facial expressions that closely resemble the GT.}
    \label{fig:express4dcm}
\end{figure}
To illustrate the advantage of semantic conditioning, we compare facial expressions generated by our method against baseline approaches. As shown in~\Cref{fig:express4dcm}, our model accurately captures the emotional context (e.g., ``Hopeless supplication'') and the fine-grained dynamic changes between keyframes. Specifically, our method successfully reflects the progressive head tilt—from ``slightly upward'' in Keyframe 1 to ``even higher'' in Keyframe 2—closely matching the GT. In contrast, the baselines struggle to interpret these nuanced text instructions. T2M-GPT and CTEG largely fail to generate the correct upward gaze and emotion, while Express4D-MDM produces nearly identical expressions for both keyframes, missing the subtle progression in the descriptions. These visual results demonstrate our model's superiority in semantic alignment.

\subsection{Ablation Study}

\paragraph{Data Augmentation.}
To verify the effectiveness of our proposed data augmentation 
strategies, we conducted fine-tuning experiments on three annotation 
types: Original, ARKit-based, and Image-based on diverse 
architectures. Consistent train-test splits were strictly maintained 
across all combinations of models and annotation types. 

\begin{table}[h]
\centering
\caption{Ablation study on data augmentation across model architectures.}
\label{tab:comprehensive_annotation_results}
\scriptsize
\setlength{\tabcolsep}{3.5pt}
\begin{tabular}{l ccc ccc}
\toprule
\multirow{2}{*}{\textbf{Model}} 
& \multicolumn{3}{c}{\textbf{MSE $\downarrow$}} 
& \multicolumn{3}{c}{\textbf{MAE $\downarrow$}} \\
\cmidrule(lr){2-4} \cmidrule(lr){5-7}
& \textbf{Original} & \textbf{Image-based} & \textbf{ARKit-based} & \textbf{Original} & \textbf{Image-based} & \textbf{ARKit-based} \\
\midrule
Qwen3-4B-Instruct-2507
& 0.01400 & \underline{0.01047} & \textbf{0.00725} 
& 0.05920 & \underline{0.05189} & \textbf{0.03926} \\
Qwen3-14B
& 0.01393 & \underline{0.01141} & \textbf{0.00691} 
& 0.05955 & \underline{0.05365} & \textbf{0.03877} \\
DeepSeek-R1-Distill-Qwen-14B
& 0.01634 & \textbf{0.01520} & \underline{0.01558} 
& 0.06614 & \underline{0.06333} & \textbf{0.06081} \\
\bottomrule
\end{tabular}
\end{table}

As shown in~\Cref{tab:comprehensive_annotation_results}, the ARKit-based annotations consistently outperform both the original and image-based versions across all evaluation metrics. 
This demonstrates that textual supervision derived from structured ARKit parameters enables the model to better learn the semantic-to-muscle correspondence of facial expressions. More detailed comparisons are included in Appendix~\ref{appendix:training}.
% Therefore, we adopt ARKit-based annotation data for the following experiments, while more detailed comparisons are included in the Appendix~\ref{appendix:training}.

\paragraph{Semantic Understanding.}
We design a rigorous ablation study to investigate the role of 
semantic information in facial parameter learning and the contribution 
of LLMs' linguistic priors. Specifically, we define two experimental 
configurations:

\begin{itemize}[leftmargin=1em]
    \item \textbf{Semantic Configuration:} This follows our standard 
    pipeline, where the model receives the full input $U_n$. Outputs 
    are formatted as structured action-value pairs 
    $\{(a_k, v_k^n)\}_{k=1}^{K}$.
    \item \textbf{Non-Semantic Configuration:} In this configuration, 
    the model receives $\tilde{U}_n$, a degraded version of $U_n$ 
    with all semantic cues removed, such as ARKit parameter 
    definitions and facial muscle mappings. Outputs are instead 
    formatted as anonymous value sets $\{v_k^n\}_{k=1}^{K}$ without 
    any parameter names.
\end{itemize}

To systematically evaluate the impact of these representations across 
different model architectures, we established four comparative 
settings using consistent train-test splits on ARKit-based annotation 
data: (i)~\textbf{Original-Non-Semantic}: direct inference with the 
frozen pre-trained model under the Non-Semantic Configuration; 
(ii)~\textbf{Original-Semantic}: direct inference with the frozen 
pre-trained model under the Semantic Configuration; 
(iii)~\textbf{LoRA-Non-Semantic}: fine-tuning on our KeyframeFace 
dataset and subsequent inference, both conducted under the 
Non-Semantic Configuration; and (iv)~\textbf{LoRA-Semantic}: 
fine-tuning on our KeyframeFace dataset and subsequent inference, 
both conducted under the Semantic Configuration.

\begin{table}[h]
\centering
\caption{Ablation study on semantic understanding and model configuration.}
\label{tab:ablation_semantic}
\scriptsize
\setlength{\tabcolsep}{6pt}
\begin{tabular}{l cc c ccc c}
\toprule
\multirow{2}{*}{\textbf{Setting}} 
& \multicolumn{2}{c}{\textbf{Coefficient Error}}
& \multicolumn{1}{c}{\textbf{Similarity}}
& \multicolumn{3}{c}{\textbf{R-Precision} $\uparrow$} 
& \multirow{2}{*}{\textbf{MMD} $\downarrow$} \\
\cmidrule(lr){2-3} \cmidrule(lr){4-4} \cmidrule(lr){5-7}
& \textbf{MSE} $\downarrow$ & \textbf{MAE} $\downarrow$ 
& \textbf{W-Dist} $\downarrow$ 
& \textbf{top-1} & \textbf{top-2} & \textbf{top-3} & \\
\midrule
\rowcolor{gray!15} \textbf{GT-test}
& - & - & 0.0055 & 0.2040 & 0.3565 & 0.4802 & 1.0600 \\
\midrule
Original-Non-Semantic-14B
& 1.5860 & 11.0991 & 1.5606 & 0.0264 & 0.0541 & 0.0865 & 1.4710 \\
Original-Non-Semantic-4B
& 0.1030 & 0.2371 & 0.1796 & 0.0325 & 0.0661 & 0.0950 & 1.4691 \\
Original-Semantic-4B
& 0.0257 & 0.0898 & 0.0585 & 0.0877 & 0.1466 & 0.2091 & 1.3220 \\
Original-Semantic-14B 
& 0.0195 & 0.0830 & 0.0192 & 0.0853 & 0.1659 & 0.2200 & 1.2947 \\
LoRA-Non-Semantic-4B
& 0.0112 & 0.0511 & 0.0310 & 0.0974 & 0.1887 & 0.2548 & 1.1779 \\
LoRA-Non-Semantic-14B
& 0.0167 & 0.0660 & 0.0323 & 0.1046 & \underline{0.1923} & 0.2572 & 1.1773 \\
LoRA-Semantic-4B
& \underline{0.0072} 
& \underline{0.0393} 
& \underline{0.0177} 
& \underline{0.1995} 
& \textbf{0.3413} 
& \underline{0.4399} 
& \underline{1.0689} \\
\textbf{LoRA-Semantic-14B}
& \textbf{0.0069}
& \textbf{0.0388} 
& \textbf{0.0145} 
& \textbf{0.2079} 
& \textbf{0.3413} 
& \textbf{0.4579} 
& \textbf{1.0602} \\
\bottomrule
\end{tabular}
\end{table}

\Cref{tab:ablation_semantic} presents results across the four experimental settings evaluated on two model scales: Qwen3-4B-Instruct-2507 and Qwen3-14B. Models trained under the Semantic Configuration significantly outperform their Non-Semantic counterparts across all evaluation metrics, particularly in coefficient-level errors (MSE and MAE), demonstrating better overall accuracy. Specifically, semantic models achieve lower W-Dist values and substantially higher R-Precision scores (nearly doubling), indicating better distributional alignment with GT data and stronger cross-modal consistency between text and motion. Remarkably, the Semantic-4B model already surpasses the Non-Semantic-14B model on most metrics, suggesting that semantic supervision contributes more to performance than model scaling alone. Furthermore, LoRA-Semantic achieves performance closest to the GT-test upper bound, validating that semantically structured representations enable LLMs to leverage linguistic priors more effectively, leading to more precise expression control and improved motion-text consistency.
% \paragraph{Motion Diffusion Model Comparison.} To compare with alternative generative paradigms, we implement the Motion Diffusion Model (MDM)\cite{tevet2022human}, a diffusion-based baseline widely adopted in motion generation tasks, including facial animation. This model is particularly relevant as it is also used in Express4D\cite{aloni2025express4d} for facial motion representation. We train this baseline on our KeyframeFace dataset using identical ARKit-based motion representations and evaluate it alongside our LLM-based approaches.

\paragraph{Input Standardization and Keyframe Count.}
We further evaluate two additional design choices. To quantify the contribution of the Input Standardization stage, we assess our Text-To-Animation model under degraded input conditions, retaining only a single component of the full structured script (scene, emotion, local expression, or colloquial description) in both full and shortened forms. Specifically, the Input Standardization stage employs Qwen3-14B to parse and normalize raw text inputs into a structured four-component script; this design ensures consistent formatting regardless of input style or verbosity. Results indicate that relying on any single component significantly increases prediction errors, confirming that the complete structured script is critical for accurate parameter generation. Additionally, we examine the effect of keyframe count by comparing model outputs generated with varying numbers of keyframes. We observe that the generation quality remains highly stable across different settings, demonstrating the model's robustness to keyframe count variations. Full results for both analyses are provided in Appendix~\ref{appendix:training}.

%% file: sec/6_conclusion.tex
\section{Conclusion}
\label{sec:conclusion}

This paper presents KeyframeFace, a language-driven framework for semantic facial animation via interpretable keyframes. 
Instead of directly regressing dense facial motion from text, our method represents animation as a sequence of semantically meaningful keyframes in an ARKit-based control space, enabling a clearer correspondence between language descriptions and facial states. 
We further construct a multimodal dataset with semantic scripts, aligned ARKit coefficients, and annotated keyframes. 
Experiments show that semantic keyframe supervision and LLM priors improve expression fidelity and semantic alignment.

\paragraph{Limitations.}
Despite these results, the framework is limited by its reliance on the ARKit blendshape space and templated scripts. 
This may restrict its ability to capture fine-grained facial dynamics and generalize to unconstrained real-world scenarios or highly complex performances. 
In addition, language-driven facial animation could be misused to generate misleading synthetic media or impersonation content, especially when combined with realistic digital humans. 
Future work will explore more expressive facial representations, broader real-world data, and interactive or real-time language-driven animation editing, while emphasizing responsible use, consent, and appropriate safeguards.

%% file: sec/appendix.tex
\section{User Study}
\label{appendix:userstudy}

\paragraph{Study Design.}
We conduct a forced-choice perceptual study to evaluate whether 
generated facial expressions are consistent with a given narrative 
script. In each trial, a participant is presented with a text script 
alongside two rendered expression outputs---one from our method and 
one from a baseline model---and is asked to select the output that 
better matches the described emotional content. Participants may also 
indicate that both outputs are similarly appropriate (\textit{similar} 
vote). Model identities are not disclosed to participants, and the 
left-right positioning of the two outputs is randomized independently 
per trial. A total of 2 evaluators participated in the study.

We report two tiers of results: (1)~\textbf{All votes}, aggregating 
all responses from both evaluators; and (2)~\textbf{Valid votes}, retaining only trials in which both evaluators 
selected the same option, providing a higher-confidence subset.

\paragraph{Results.}
\Cref{tab:userstudy_full} reports the complete results.
Our method achieves a win rate above 63\% across all three baselines 
under the all-votes condition.
The advantage is most pronounced against Express4D-MDM and T2M-GPT 
(79.2\% and 78.9\%, respectively), while the margin against CTEG is 
smaller (63.4\%), consistent with CTEG being the strongest baseline 
in quantitative metrics as well.
Under the valid-votes condition, win rates increase substantially to 
89.7\%, 72.4\%, and 91.7\% against Express4D-MDM, CTEG, and T2M-GPT, 
respectively, indicating that the preference for our method is 
consistent across both evaluators rather than driven by individual 
variation.

\begin{table}[h]
\centering
\caption{Full user study results. }
\label{tab:userstudy_full}
\footnotesize
\setlength{\tabcolsep}{4pt}
\begin{tabular}{ll rrr c}
\toprule
\textbf{Baseline} & \textbf{Votes} 
& \textbf{Win} & \textbf{Lose} & \textbf{Sim}
& \textbf{Win Rate} $\uparrow$ \\
\midrule
\multirow{2}{*}{Express4D-MDM}
  & All    & 84  & 22 & 94  & 79.2\% \\
  & Valid  & 35  &  4 & 42  & 89.7\% \\
\midrule
\multirow{2}{*}{CTEG}
  & All    & 116 & 67 & 17  & 63.4\% \\
  & Valid  & 42  & 16 &  3  & 72.4\% \\
\midrule
\multirow{2}{*}{T2M-GPT}
  & All    & 146 & 39 & 15  & 78.9\% \\
  & Valid  & 55  &  5 &  0  & 91.7\% \\
\bottomrule
\end{tabular}
\end{table}

\section{Training and Implementation Details}
\label{appendix:training}
\subsection{Extensive Model Experimentation}

We conducted a comprehensive performance evaluation on KeyframeFace Data using three mainstream LLMs: Qwen3-4B-Instruct-2507, Qwen3-14B, and DeepSeek-R1-Distill-Qwen-14B. Experiments span three annotation types 
(ARKit-based, Image-based, Original) and two configurations 
(semantic and non-semantic), under both fine-tuned (LoRA) and frozen 
(Original) settings. All 36 configurations employ consistent train-test 
splits to ensure fair comparison.

To systematically identify each experimental setup, we establish a rigorous naming convention formatted as \textbf{Tuning}-\textbf{Model}-\textbf{Annotation}-\textbf{Configuration}, where:
\begin{itemize}
    \item \textbf{Tuning} indicates the model's training state (\textit{LoRA} for fine-tuning, or \textit{Original} for frozen pre-trained);
    \item \textbf{Model} specifies the underlying LLM architecture;
    \item \textbf{Annotation} defines the data augmentation type (\textit{ARKit}, \textit{Image}, or \textit{Original});
    \item \textbf{Configuration} denotes whether the \textit{semantic} or \textit{non-semantic} input configuration is applied.
\end{itemize}
\begin{table}[b!]
\centering
\caption{Loss Comparison Across Different Configurations}
\label{tab:training_loss}
\scriptsize
\setlength{\tabcolsep}{3pt}
\begin{tabular}{llcc}
\toprule
Model & Data Type & Train Loss & Eval Loss \\
\midrule
\multirow{6}{*}{\makecell[l]{\textbf{Qwen3-4B-}\\\textbf{Instruct-2507}}} 
& Original-semantic & \underline{0.04399845} & 0.15115031 \\
& ARKit-semantic & \textbf{0.04079014} & \textbf{0.04607984} \\
& Image-semantic & 0.04674645 & \underline{0.05088831} \\
& Original-non-semantic & 0.11204951 & 0.12501122 \\
& ARKit-non-semantic & 0.10009367 & 0.11942095 \\
& Image-non-semantic & 0.10147048 & 0.12050562 \\
\midrule
\multirow{6}{*}{\textbf{Qwen3-14B}} 
& Original-semantic & 0.04413972 & 0.05011869 \\
& ARKit-semantic & \textbf{0.03702557} & \textbf{0.04481282} \\
& Image-semantic & \underline{0.04356491} & \underline{0.04925121} \\
& Original-non-semantic & 0.10224197 & 0.12059103 \\
& ARKit-non-semantic & 0.09825735 & 0.11423669 \\
& Image-non-semantic & 0.09463592 & 0.11857172 \\
\midrule
\multirow{6}{*}{\makecell[l]{\textbf{DeepSeek-R1-}\\\textbf{Distill-Qwen-14B}}} 
& Original-semantic& 0.04506757 & 0.05090072 \\
& ARKit-semantic& \textbf{0.03918483} & \textbf{0.04624721} \\
& Image-semantic& \underline{0.04335775} & \underline{0.04956503} \\
& Original-non-semantic & 0.10576349 & 0.12249377 \\
& ARKit-non-semantic & 0.09627808 & 0.11671870 \\
& Image-non-semantic & 0.09928249 & 0.12036505 \\
\bottomrule
\end{tabular}
\end{table}

\paragraph{Semantic Representation and Annotation Type.}
As shown in~\Cref{tab:mae_results_desc}, semantic representation 
plays a decisive role: non-semantic format causes severe performance 
degradation in frozen models, while ARKit-based annotations consistently 
achieve the best reconstruction accuracy across all settings. Only 
\textbf{LoRA-qwen3-14b-ARKit-semantic} and 
\textbf{LoRA-qwen3-4b-ARKit-semantic} reach breakthrough MAE levels 
around 0.039, demonstrating the synergistic effect of ARKit annotations 
and semantic input. Additionally, Table~\ref{tab:training_loss} shows 
that ARKit-based annotations consistently yield the lowest training and 
evaluation loss across all three model architectures, further supporting 
their adoption as the primary annotation type in our main experiments.

\paragraph{Training and Evaluation Loss.}
Table~\ref{tab:training_loss} further provides complete training and 
evaluation loss across all annotation types and representation formats, 
complementing the main paper's 
Table~\ref{tab:comprehensive_annotation_results} which reports only 
semantic configurations. Notably, non-semantic formats consistently 
incur significantly higher loss (around 0.10--0.12) compared to their 
semantic counterparts (around 0.04), further confirming the critical 
role of semantic guidance in model optimization.

\begin{table}[t]
\centering
\caption{Model Performance Evaluation Results (Sorted by MAE).}
\label{tab:mae_results_desc}
\scriptsize
\begin{tabular}{lcc}
\toprule
Model Configuration & Overall MSE $\downarrow$& Overall MAE $\downarrow$\\
\midrule
Original-deepseek-ARKit-non-semantic & 198.39302 & 3.345304 \\
Original-qwen3-14b-Image-non-semantic & 8.532402 & 2.218955 \\
Original-qwen3-14b-Original-non-semantic & 7.814500 & 2.061766 \\
Original-deepseek-Original-non-semantic & 77.067899 & 1.663504 \\
Original-qwen3-14b-ARKit-non-semantic & 11.099144 & 1.585984 \\
LoRA-deepseek-ARKit-non-semantic & 35.721793 & 0.904577 \\
LoRA-deepseek-Image-non-semantic & 13.330677 & 0.414620 \\
LoRA-deepseek-Original-non-semantic & 12.071420& 0.387600\\
Original-qwen3-4b-Image-non-semantic & 0.123285 & 0.261439 \\
Original-qwen3-4b-ARKit-non-semantic & 0.103033 & 0.237061 \\
Original-qwen3-4b-Original-non-semantic & 0.085073 & 0.213044 \\
Original-qwen3-4b-Image-semantic & 0.033737 & 0.093828 \\
Original-qwen3-4b-ARKit-semantic & 0.025687 & 0.089817 \\
Original-qwen3-14b-Image-semantic & 0.027363 & 0.088729 \\
Original-deepseek-Image-semantic & 0.024913 & 0.086483 \\
Original-qwen3-4b-Original-semantic & 0.024299 & 0.083361 \\
Original-deepseek-ARKit-semantic & 0.023788 & 0.083278 \\
Original-qwen3-14b-ARKit-semantic & 0.019482 & 0.082952 \\
Original-deepseek-Original-semantic & 0.022272 & 0.082861 \\
Original-qwen3-14b-Original-semantic & 0.022450 & 0.081556 \\
LoRA-qwen3-14b-Image-non-semantic & 0.017503 & 0.068929 \\
LoRA-qwen3-14b-Original-non-semantic & 0.017331 & 0.066917 \\
LoRA-deepseek-Original-semantic & 0.016335 & 0.066142 \\
LoRA-qwen3-14b-ARKit-non-semantic & 0.016676 & 0.065985 \\
LoRA-qwen3-4b-Original-non-semantic & 0.016216 & 0.065823 \\
LoRA-deepseek-Image-semantic & 0.015204 & 0.063333 \\
LoRA-deepseek-ARKit-semantic & 0.015578 & 0.060807 \\
LoRA-qwen3-14b-Original-semantic & 0.013933 & 0.059548 \\
LoRA-qwen3-4b-Original-semantic & 0.013998 & 0.059197 \\
LoRA-qwen3-4b-Image-non-semantic & 0.012006 & 0.055452 \\
LoRA-qwen3-14b-Image-semantic & 0.011411 & 0.053649 \\
LoRA-qwen3-4b-Image-semantic & 0.010472 & 0.051885 \\
LoRA-qwen3-4b-ARKit-non-semantic & 0.011215 & 0.051067 \\
LoRA-qwen3-4b-ARKit-semantic & 0.007245 & 0.039264 \\
LoRA-qwen3-14b-ARKit-semantic & \textbf{0.006906} & \textbf{0.038765} \\
\bottomrule
\end{tabular}
\end{table}

\begin{algorithm}[!t]
\caption{KeyframeFace Generation Procedure}
\label{alg:generation}
\KwIn{User input $\mathcal{T}$; System Prompt $P$}
\KwOut{ARKit Keyframe Sequence $\mathcal{O}$}
\textbf{Stage I: Input Standardization}\\
Standardize $\mathcal{T}$ into Script $S$: $S \leftarrow \text{LLM}(\mathcal{T})$\;
\BlankLine
\textbf{Stage II: Text-To-Animation}\\
Extract keyframe descriptions from $S$: $\{(F_n, n)\}_{n=1}^{N} \leftarrow f_{\text{key}}(S)$\;
\ForEach{$(F_n, n) \in \{(F_n, n)\}_{n=1}^{N}$}{
Derive Target Frame Instruction: $I_{n} \leftarrow f_{\text{task}}(F_n, n)$\;
Compose Full Input: $U_n \leftarrow P \oplus S \oplus I_{n}$\;
$\{(a_k, v_k^n)\}_{k=1}^{K} \leftarrow \text{LLM}_{\text{ft}}(U_n)$\;
Append $\{(a_k, v_k^n)\}_{k=1}^{K}$ to $\mathcal{O}$\;
}
\Return{$\mathcal{O}$}\;
\end{algorithm}

\subsection{Input Standardization Ablation Study}
\label{appendix:is_keyframe_ablation}

To examine the contribution of the Input Standardization stage, we 
construct eight degraded input variants by retaining only a single 
component of the full structured script (scene, emotion, local 
expression, or colloquial description), each in full and shortened 
forms, and compare against our full pipeline. 
Table~\ref{tab:is_ablation_descending} reports the results. All degraded 
variants exhibit substantially higher error than the full pipeline, demonstrating that the 
complete structured script produced by the Input Standardization 
stage is critical for accurate ARKit parameter generation. 

\begin{table}[h]
\centering
\caption{Input Standardization Ablation Study.}
\label{tab:is_ablation_descending}
\scriptsize
\setlength{\tabcolsep}{5pt}
\begin{tabular}{lcc}
\toprule
\textbf{Input Condition} & \textbf{MSE} $\downarrow$ & \textbf{MAE} $\downarrow$ \\
\midrule
Local (short)           & 0.1092             & 0.1194             \\
Colloquial (short)      & 0.1079             & 0.1178             \\
Colloquial              & 0.1080             & 0.1176             \\
Local                   & 0.1084             & 0.1172             \\
Scene (short)           & 0.1073             & 0.1157             \\
Emotion (short)         & 0.1073             & 0.1156             \\
Emotion                 & \underline{0.1072} & 0.1152             \\
Scene                   & \underline{0.1072} & \underline{0.1148} \\
\midrule
\rowcolor{gray!15} KeyframeFace & \textbf{0.0069} & \textbf{0.0388} \\
\bottomrule
\end{tabular}
\end{table}

\subsection{Keyframe Count Ablation Study}
We compare model outputs generated with different keyframes to 
examine the effect of keyframe count on generation quality. 
As shown in~\Cref{tab:keyframe_ablation}, performance remains 
stable across all three settings, suggesting that generation quality 
is robust to the number of keyframes.

\begin{table}[h]
\centering
\caption{Keyframe Count Ablation Study.}
\label{tab:keyframe_ablation}
\scriptsize
\setlength{\tabcolsep}{5pt}
\begin{tabular}{ccc} % 修改此处，将第一个 l 改为 c
\toprule
\textbf{Keyframe Count} & \textbf{MSE} $\downarrow$ & \textbf{MAE} $\downarrow$ \\
\midrule
1 & \textbf{0.0064} & \underline{0.0383} \\
2 & \underline{0.0066} & \textbf{0.0374} \\
3 & 0.0073          & 0.0403          \\
\bottomrule
\end{tabular}
\end{table}

\subsection{Comparison Results Discussion}
\label{app:baseline_analysis}

\paragraph{Express4D-MDM.} Express4D-MDM shows significantly lower 
R-Precision, reflecting a fundamental limitation of diffusion-based 
approaches in this domain: the model treats text embeddings and ARKit 
coefficients as opaque numerical distributions to be aligned 
statistically, without leveraging the inherent semantic structure on 
either side. In contrast, our LLM-based approach benefits from 
semantic grounding at both ends---the LLM comprehends the linguistic 
intent of the input description, while ARKit coefficients explicitly 
correspond to named, anatomically grounded facial actions. This 
dual semantic alignment allows our method to directly bridge 
linguistic meaning and physical facial motion, a correspondence 
that purely numerical diffusion objectives cannot capture.

\paragraph{T2M-GPT.} T2M-GPT exhibits a significant performance drop when adapted to text-driven facial expression generation, struggling to achieve high R-Precision and showing higher coefficient errors. We observe that this degradation primarily stems from the architectural bias of autoregressive next-token prediction over discrete codebooks. Unlike full-body motion, facial dynamics are characterized by subtle, highly semantic localized movements. We found that the autoregressive paradigm tends to suffer from prior collapse in this domain, heavily favoring a narrow set of high-frequency neutral codes while neglecting the fine-grained text conditioning. In contrast, our KeyframeFace avoids this temporal error accumulation and discrete quantization loss by explicitly mapping text prompts to semantic keyframes, allowing for precise emotional and contextual alignment without sacrificing generation diversity.

\paragraph{CTEG.} As a sequence generation model relying on standard regression objectives for frame-level prediction, CTEG is fundamentally constrained by the inherent "one-to-many" ambiguity of text-driven animations—where a single prompt can validly correspond to diverse expressions. To minimize the regression loss under such uncertainty, the model's optimal strategy converges to predicting the weighted average of all possible outcomes. This leads to a severe "regression to the mean" effect, resulting in overly smoothed and compressed facial dynamics. Ultimately, this deterministic averaging prevents the model from precisely localizing and rendering fine-grained, detailed keyframes.

\section{Details of the ARKit-To-Visual Animation Module}
\label{appendix:details}
\subsection{Duration Predictor}
Given two adjacent keyframe vectors $\mathbf{v}_n, \mathbf{v}_{n+1} \in \mathbb{R}^K$, the duration predictor $\mathcal{D}$ estimates the intrinsic number of intermediate frames:$$\hat{M} = \mathcal{D}([\mathbf{v}_n; \mathbf{v}_{n+1}])$$where $[\cdot;\cdot]$ denotes vector concatenation. $\mathcal{D}$ is implemented as a three-layer MLP. To train the predictor, we extract the number of intermediate frames between different keyframes from the KeyframeFace Dataset as the ground-truth $M^*$, and train the model using MSE loss:$$\mathcal{L}_{\mathcal{D}} = \|\hat{M} - M^*\|^2$$To support flexible motion control, the predicted frame count is modulated by a user-specified speed scale factor $\alpha \in \mathbb{R}^+$:$$M = \lfloor \hat{M} \cdot \alpha \rceil$$where $\alpha < 1$ produces faster transitions, $\alpha > 1$ produces slower transitions, and $\alpha = 1$ preserves the naturally predicted duration.
\subsection{Motion Interpolator}
\paragraph{Overview.} The motion interpolator $\mathcal{F}$ integrates boundary keyframe contexts with temporal query vectors, processes them through Transformer blocks via cross- and self-attention to ensure temporal coherence, and finally projects the refined representations back to the ARKit parameter space to yield the intermediate sequence.
\paragraph{Encoding.} The two keyframes are independently projected into a $d=128$-dimensional hidden space:$$\mathbf{k}_n = \mathbf{W}_s \mathbf{v}_n, \quad \mathbf{k}_{n+1} = \mathbf{W}_e \mathbf{v}_{n+1}, \quad \mathbf{k}_n, \mathbf{k}_{n+1} \in \mathbb{R}^d$$.For the $q$-th intermediate frame ($q \in \{1,\ldots,Q\}$), its normalized temporal position is defined as $\tau_q = \frac{q}{Q+1} \in (0, 1)$. The corresponding query vector is computed via sinusoidal positional encoding:$$\mathbf{q}_q = \text{PE}(\tau_q) = \left[\sin(w_1\tau_q), \cos(w_1\tau_q), \ldots, \sin(w_{d/2}\tau_q), \cos(w_{d/2}\tau_q)\right] \in \mathbb{R}^d$$where $\{w_j\}_{j=1}^{d/2}$ are predefined frequencies of increasing magnitude, encoding both coarse and fine-grained temporal positions.

\paragraph{Transformer Layers.}
The model consists of $L=4$ stacked Transformer blocks, each applying 
cross-attention followed by self-attention with residual connections 
and layer normalization. In cross-attention, each query attends to the 
keyframe context $\{\mathbf{k}_n, \mathbf{k}_{n+1}\}$, naturally 
reflecting temporal proximity. Self-attention then enforces global 
temporal coherence across all $Q$ intermediate frames jointly.

\paragraph{Output Projection.}
After $L$ Transformer blocks, the refined intermediate frame 
representations are projected back to the ARKit parameter space 
via a linear layer, yielding the complete intermediate sequence 
$\{\hat{\mathbf{v}}_q\}_{q=1}^Q \subset \mathbb{R}^{61}$.

\paragraph{Training Objective.}
The model is supervised with a composite loss:
$$\mathcal{L} = \mathcal{L}_{\text{recon}} + \lambda_1 
\mathcal{L}_{\text{vel}} + \lambda_2 \mathcal{L}_{\text{acc}}$$
where $\mathcal{L}_{\text{recon}}$ measures per-frame reconstruction 
error, $\mathcal{L}_{\text{vel}}$ penalizes first-order temporal 
inconsistencies to encourage smooth transitions, and 
$\mathcal{L}_{\text{acc}}$ constrains second-order differences to 
suppress unnatural abrupt motion changes. $\lambda_1 = 0.1$ and 
$\lambda_2 = 0.01$ are empirically set weighting coefficients.

\paragraph{Training Data.} Training samples are constructed from the KeyframeFace dataset, which comprises continuously captured facial motion sequences recorded at a fixed frame rate. Each sequence is segmented into clips of fixed length $S=50$, where the first frame $\mathbf{v}_n$ and the last frame $\mathbf{v}_{n+1}$ serve as boundary keyframes, and the $Q = S - 2 = 48$ intermediate frames provide ground-truth supervision. This formulation ensures that the model learns to recover realistic, data-driven motion trajectories rather than relying on heuristic interpolation assumptions.

\subsection{Inference} At inference time, the Motion Interpolator $\mathcal{F}$ is conditioned on the boundary keyframes $\mathbf{v}_n$ and $\mathbf{v}_{n+1}$, and generates a fixed-length intermediate sequence of $Q=48$ frames. To reconcile this fixed output length with the target frame count $M$ determined by the Duration Predictor, the generated sequence is temporally resampled via linear interpolation along the time axis:$$\hat{\mathbf{v}}'_m = \hat{\mathbf{v}}_{\lfloor m \cdot \frac{Q}{M} \rfloor} \cdot \left(1 - r_m\right) + \hat{\mathbf{v}}_{\lceil m \cdot \frac{Q}{M} \rceil} \cdot r_m, \quad m \in \{1, \ldots, M\}$$where $r_m = m \cdot \frac{Q}{M} - \lfloor m \cdot \frac{Q}{M} \rfloor$ is the fractional remainder. The resampled sequence $\{\hat{\mathbf{v}}'_m\}_{m=1}^{M}$ is then concatenated with the original keyframes to construct the complete ARKit parameter stream:$$\mathcal{O}_{n} = [\mathbf{v}_n,\ \hat{\mathbf{v}}'_1,\ \hat{\mathbf{v}}'_2,\ \ldots,\ \hat{\mathbf{v}}'_M,\ \mathbf{v}_{n+1}]$$This process is applied to each consecutive keyframe pair, and the resulting segments are concatenated to form the full animation sequence $\mathcal{O}$, which is subsequently rendered into high-fidelity visual output through the MetaHuman system.

\section{Evaluation Model Details}
\label{appendix:rprecision_mmd_training}

\subsection{Win Rate}
\label{appendix:win_rate}
The user study assesses the perceptual quality and script-consistency of the generated facial animations. To quantify user preferences, we report the Win Rate, defined as:
\begin{equation}
\mathrm{Win\ Rate} = \frac{W}{W + L},
\end{equation}
where $W$ and $L$ denote the number of trials in which our method is preferred over (wins) and loses to the baseline, respectively. Trials in which the participant deems the two results as \textit{similar} are strictly excluded from both the numerator and denominator.

\subsection{R-Precision}
R-Precision is used to evaluate retrieval performance by examining whether the ground-truth motion corresponding to a given text description appears within the top-$K$ ranked candidates, where $K=3$. For a batch of size $N=16$, we compute a cosine-similarity matrix $S_{ij}=\mathrm{sim}(\mathcal{T}_i, A_j)$ and determine the rank of the matched motion $A_i$ for each query $\mathcal{T}_i$. R-Precision@$K$ is then defined as
\begin{equation}
\mathrm{R\text{-}Precision@K}
=
\frac{1}{N}
\sum_{i=1}^{N}
\mathbf{1}
\left[
\operatorname{rank}(A_i \mid \mathcal{T}_i) \le K
\right],
\end{equation}
indicating the proportion of queries whose correct matches fall within the top-$K$. Higher values reflect stronger cross-modal discriminative ability.

\subsection{Multimodal Distance (MMD)}
Multimodal Distance measures the geometric proximity between matched text and motion embeddings in the learned joint space. For each sample pair $(\mathcal{T}_i, A_i)$, we compute the Euclidean distance between their normalized embeddings and take the average over all pairs:
\begin{equation}
\mathrm{MMD}
=
\frac{1}{N}
\sum_{i=1}^{N}
\left\|
f_{\mathrm{text}}(\mathcal{T}_i)
-
f_{\mathrm{motion}}(A_i)
\right\|_2 .
\end{equation}
Lower values indicate that the text and motion embeddings are located closer together, suggesting stronger semantic alignment.

\subsection{Evaluation Model Training Details}
The evaluation model is trained end-to-end using a symmetric InfoNCE loss, which encourages matched text--motion pairs to have high similarity while pushing unmatched samples apart. The loss is defined as
\begin{equation}
\mathcal{L}
=
\frac{1}{2}
\left(
\mathcal{L}_{\mathcal{T}\rightarrow A}
+
\mathcal{L}_{A\rightarrow \mathcal{T}}
\right),
\end{equation}
with a temperature parameter $\tau=0.07$, and is applied over all text--motion pairs within each batch. Training employs the AdamW optimizer with a learning rate of $1\times 10^{-4}$ and weight decay of the same magnitude, together with a cosine-annealing learning-rate schedule. We train the model for up to 1000 epochs and apply early stopping when the R-Precision@1 on the validation split does not improve for ten consecutive epochs. Both the text encoder and motion encoder are fully trainable during optimization. Motion features are standardized using Z-score statistics computed from the training set, while text descriptions are tokenized with the MiniLM tokenizer. A fixed random seed of 42 is used throughout to ensure reproducibility.

\section{KeyframeFace Dataset Details}
\label{appendix:data_v}

\subsection{Data Augmentation}
\label{subsec:data_augmentation_app}
While the initial keyframe descriptions capture high-level semantic intent, professional actors often introduce subtle, context-specific facial nuances beyond the written script. To comprehensively capture these fine-grained details, we augment each keyframe’s annotation using both LLMs and MLLMs.

\paragraph{ARKit-Based Annotation.}
% \noindent\textbf{ARKit-Based Annotation.} 
We enrich the annotation by converting ARKit coefficients into textual facial descriptions leveraging Qwen3-30B-A3B-Instruct\cite{yang2025qwen3}. 
Given that ARKit parameters directly correspond to interpretable facial muscle activations, providing specific blendshape names and their respective values enables the model to infer anatomically grounded expressions.
Our prompt defines the model’s role as a professional facial-expression expert proficient in the ARKit system, instructing it to generate concise keyframe descriptions. Notably, the model is tasked with synthesizing natural descriptions without explicitly referencing raw parameter names or values. 
For example, given inputs such as ``JawOpen:0.8, EyeBlinkLeft:1.0...'', the model might generate: 
\begin{quote}
\small
``\textit{The subject's mouth is opened widely, suggesting surprise, while the left eye is closed in a blink ...}''
\end{quote}
This procedure yields muscle-level granularity that complements the macro-level emotional intent found in the original scripts, thereby establishing a precise and anatomically grounded linguistic layer for each keyframe.

% \noindent\textbf{Image-Based Annotation.}
% \vspace{-0.8em}
\paragraph{Image-Based Annotation.}To establish a visual anchor orthogonal to the numerical ARKit parameters, we leverage the multimodal capabilities of Qwen3-VL-30B-A3B-Instruct\cite{yang2025qwen3} to synthesize descriptions directly from keyframe images. This modality captures perceptual nuances—such as transient skin deformations, periorbital tension, and the deepening of nasolabial folds—that are not explicitly encoded in the $61$-dimensional blendshape geometry. 
By prompting the model to operate as a facial-expression expert, we extract detailed descriptions of fine-grained facial features and muscle dynamics based purely on visual evidence.
In complex scenarios, such as an ``on-the-spot defense''  featuring a ``\textit{hesitant} $\rightarrow$ \textit{confident}'' transition, the model effectively captures subtle cues including gaze shifts, brow symmetry, and the relaxation of cheek muscles.
These image-derived descriptions reveal micro-level variations that are often absent from the original scripts.

% \vspace{-0.8em}
\paragraph{Quality Control and Data Integrity.}To ensure rigorous data integrity and prevent potential semantic leakage, we implemented a strict validation protocol for all augmented annotations. During generation, the models received only raw ARKit coefficients (and their technical definitions) or keyframes, with all situational context withheld to ensure descriptions were grounded in physical motion. Annotators manually reviewed all generated descriptions against the original videos, verifying that the text faithfully reflects the actual facial performance while strictly avoiding explicit ARKit coefficient values or script-specific terminology. This rigorous process confirms that all generated descriptions are grounded in physically captured ARKit coefficients and real actor performances, rather than model-internal priors. As annotation generation and model training are entirely decoupled 
stages with human-in-the-loop validation, the pipeline remains open-loop and physically grounded throughout.

\subsection{Keyframe Density and Structural Patterns} In facial expression analysis, the number of keyframes directly affects the granularity of emotional expression and model training effectiveness. As shown in Figure~\ref{fig:dataset_stats}(c), our dataset exhibits a reasonable keyframe distribution pattern: all 2,100 video clips contain at least one keyframe, with the vast majority of clips containing two keyframes, and a significant portion containing three keyframes, while clips containing four or five keyframes are relatively rare. This distribution reflects that most emotional expression clips have clear onset-climax or onset-development-climax structural patterns, while a few complex clips exhibit richer multi-stage emotional changes.
Notably, since our method focuses on generating ARKit parameters for individual keyframes based on text descriptions, the model actually learns the mapping relationship from text semantics to facial expressions at specific moments, rather than relying on sequence length-dependent temporal modeling. This design philosophy gives our method unique advantages: even under the training distribution dominated by 1--3 frames, the model can effectively generalize to complex emotional expression tasks containing more keyframes, demonstrating excellent cross-sequence-length generalization capability. This natural distribution of keyframe numbers not only reflects the characteristics of real emotional expressions but also provides ideal experimental conditions for validating the generalization and robustness of our method.

\subsection{Actor Contribution and Data Balance} As shown in Figure~\ref{fig:dataset_stats}(a), each actor contributed a roughly balanced number of clips (100 clips per actor), though total recording durations vary somewhat. This design intentionally preserves the individualized characteristics of different actors, including unique facial structures, performance habits, and emotional expression styles, providing diverse training samples for model learning. Importantly, since ARKit blendshape coefficients encode muscle activation magnitudes independent of facial geometry, the learned mapping targets the \textbf{expressive motion space} rather than the \textbf{identity space}. Generalization to unseen identities is therefore an orthogonal concern: predicted coefficients can be retargeted to any ARKit-compatible rig as a standard post-processing step. The 21 actors thus provide coverage of semantically diverse expression dynamics, while the 2,100 scripts ensure broad coverage of the linguistic-to-motion mapping space.

\subsection{Temporal Characteristics of Video Clips} 
\begin{figure}[t]
  \centering
  \includegraphics[width=\linewidth]{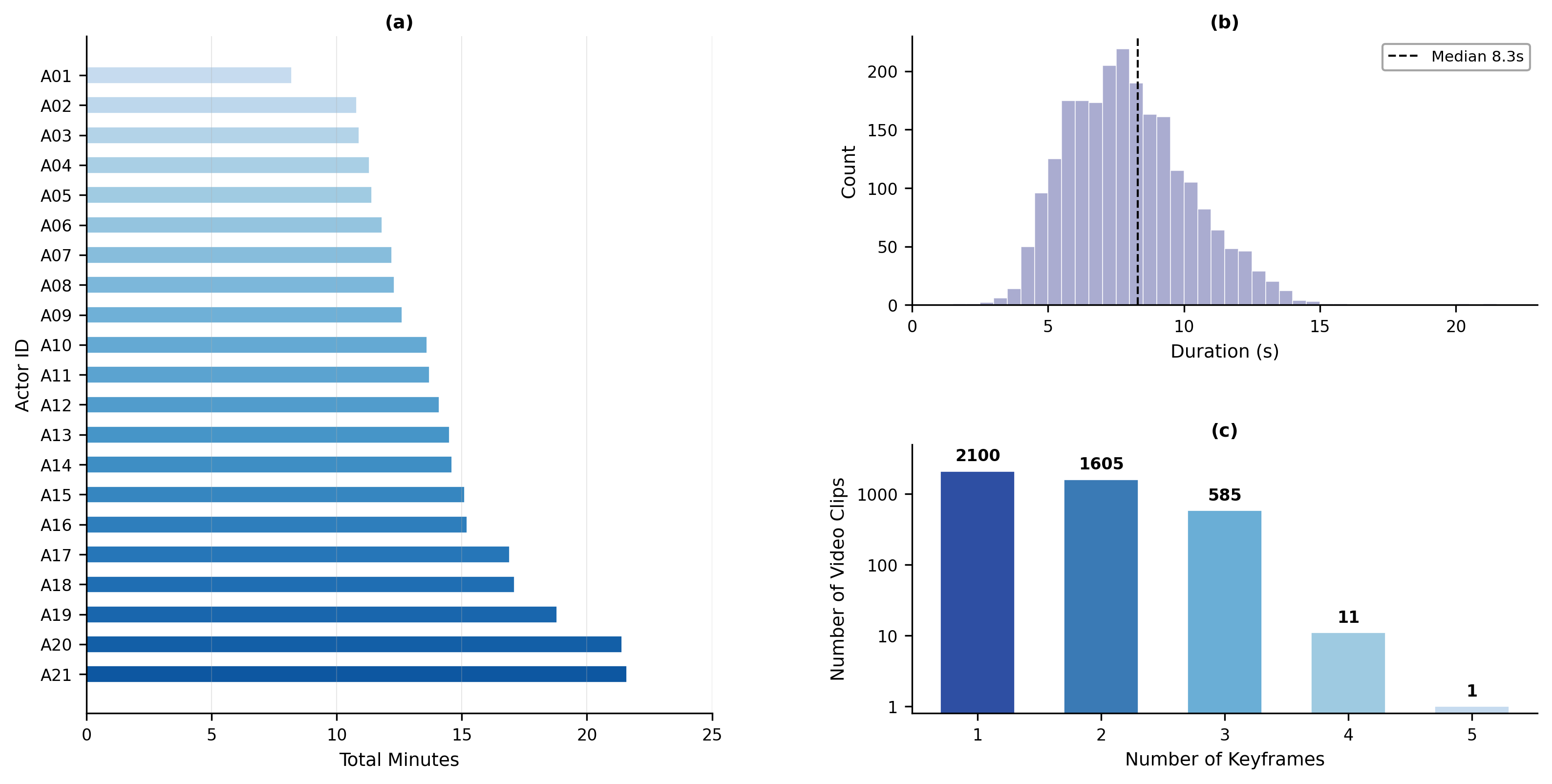}
  \caption{\textbf{KeyframeFace Dataset Statistics.}
  (a) Total recording duration per actor ($\mathrm{A}01$–$\mathrm{A}21$), sorted in ascending order.
  (b) Distribution of video clip durations, with a median of 8.3 seconds.
  (c) Distribution of keyframe counts per clip on a logarithmic scale,
  showing that most clips contain 1--3 keyframes.}
  \label{fig:dataset_stats}
\end{figure}

The video clip duration distribution in our dataset is reasonable and practical. As shown in Figure~\ref{fig:dataset_stats}(b), clip durations range from 1.9 to 21.3 seconds with a median of 8.3 seconds. Short clips mainly capture instantaneous emotional reactions (such as surprise, fear), while long clips contain complete emotional development and change processes. Notably, 75\% of clips are concentrated within the relatively narrow range of 6.6-10.3 seconds, and this consistency greatly simplifies batch processing operations and frame sequence sampling strategies during model training.

\begin{figure*}[t]
\centering
\includegraphics[width=\linewidth]{picture/dataset2.jpg}
\caption{\textbf{Emotion Distribution and Visualization.} (a) t-SNE projection of keyframes colored by emotion intensities. (b) Distribution of dominant emotions across all keyframes.}
\label{fig:emotion_distribution}
\end{figure*}

\subsection{Frame-level Emotion Analysis}

To comprehensively characterize the emotional dynamics within our dataset, we conduct fine-grained emotion analysis at the keyframe level. Since character emotions evolve with narrative development in scripts, single video-level labels are insufficient to capture this dynamic variation. Therefore, we extract keyframes from each video and employ the Qwen3-30B model for quantitative emotion assessment.

Following Ekman's emotion theory framework, we define six basic emotion categories: Sadness, Surprise, Anger, Happiness, Disgust, and Fear. For each keyframe, the model comprehensively analyzes ARKit facial expression parameters (52 blendshape parameters and 9 head pose parameters), script contextual descriptions, and emotional context to generate intensity scores ranging from 0 to 1 for each of the six emotions, ultimately outputting structured JSON-formatted emotion parameters. This keyframe-based multimodal analysis approach enables precise capture of subtle emotional expressions that vary over time in facial animations.

As shown in Figure~\ref{fig:emotion_distribution}(b), the dataset is dominated by sadness (37.7\%) and happiness (36.8\%), which together comprise nearly 75\% of keyframes, reflecting typical dramatic emotional contrast. The remaining emotions—surprise (7.9\%), anger (7.7\%), fear (7.3\%), and disgust (2.6\%)—appear less frequently but are essential for representing a full emotional spectrum.

Figure~\ref{fig:emotion_distribution}(a) visualizes the t-SNE projection of six-dimensional emotion vectors. Distinct clusters—for example, happiness and sadness—indicate clear emotional extremes, while smooth color transitions reveal abundant mixed and intermediate emotions. This demonstrates that the dataset spans both pure and composite emotional states.

In summary, our dataset demonstrates broad coverage and realistic distribution characteristics across three dimensions: emotion categories, intensity ranges, and emotion combinations. This provides high-quality, diverse training data for downstream tasks such as facial expression generation and emotion recognition, offering significant research value and application potential.